\documentclass[10pt,twocolumn,letterpaper]{article}

\usepackage{cvpr}
\usepackage{times}
\usepackage{epsfig}
\usepackage{graphicx}
\usepackage{amsmath}
\usepackage{amssymb}
\usepackage{physics}
\usepackage{bm}
\usepackage{bbold}
\usepackage{xcolor}

\usepackage[breaklinks=true,bookmarks=false]{hyperref}

\cvprfinalcopy 

\def\cvprPaperID{****} 

\begin{document}

\title{Random Relabeling for Efficient Machine Unlearning}

\author{Junde Li and Swaroop Ghosh\\
Department of Computer Science and Engineering\\
Pennsylvania State University,
University Park, PA 16802\\
{\tt\small \{jul1512, szg212\}@psu.edu}
}

\maketitle

\begin{abstract}
  Learning algorithms and data are the driving forces for machine learning to bring about tremendous transformation of industrial intelligence. However, individuals' right to retract their personal data and relevant data privacy regulations pose great challenges to machine learning: how to design an efficient mechanism to support certified data removals. Removal of previously seen data known as \textit{machine unlearning} is challenging as these data points were implicitly memorized in training process of learning algorithms. Retraining remaining data from scratch straightforwardly serves such deletion requests, however, this \textit{naive} method is not often computationally feasible. We propose the unlearning scheme \textit{random relabeling}, which is applicable to generic supervised learning algorithms, to efficiently deal with sequential data removal requests in the online setting. A less constraining removal certification method based on probability distribution similarity with \textit{naive} unlearning is further developed for logit-based classifiers.

\end{abstract}

\section{Introduction}
Valuable insights drawn from data-driven machine learning improve many aspects of human society to an increasingly intelligent level. While individuals may decide when their personal data can be used, their requests for data deletion from dataset should always be respected. The data deletion right starts being protected by recently enacted regulations such as the European Union's General Data Protection Regulation - right to be forgotten \cite{gdpr}, and United States' Consumer Data Privacy and Security Act of 2020 - right to erasure \cite{dpsa}. The regulations have broad impact on technology companies that use personal data from these regions, because continued use of the data that has been requested for removal is considered illegal. Therefore, designing an mechanism that supports efficient data deletion from trained machine learning models becomes inevitable for the affected technology companies, especially in the face of increasing individual awareness of data privacy. How can these companies deal with data usage consent withdraws?

The \textit{naive} approach to data deletion that comes readily to mind is to retrain remaining data points from scratch. This method is only computationally practical for ML models trained with small-scale dataset. However, this is not the case for many current models where the training might take days or even longer. And sequential removal requests from data owners exacerbates the efficiency issue. The problem of efficiently erasing data points is challenging as ML models unintentionally memorize previously seen data points \cite{carlini}. Authors in \cite{villaronga} claimed it might be impossible to fulfill the right to be forgotten of former  \cite{gdpr} regulation for ML models. Aside from deletion efficiency, the maximum divergence between the model unlearned from some instances and the model retrained on remaining dataset excluding these instances has to be bounded to guarantee unsuccessful \textit{membership inference} attacks \cite{carlini, yeom} for removed data points.

We only focus on the specific setting of logit-based classifiers, though our \textit{random relabeling} unlearning mechanism may be applicable to other types of supervised learning algorithms. The problem of data deletion from a statistical model is termed as \textit{machine unlearning}. Suppose a $k$-class classifier is trained on $n$ data points. Unlearning of $m$ sequential instances removes the influence of these data points on the model, which makes the unlearned model approximate enough to the model retrained with only $n-m$ data points. Machine unlearning quality, i.e. the divergence between unlearned model and retrained model, is measured on the basis of membership inference. And \textit{certified unlearning} refers to the data deletion mechanism such that adversary cannot extract membership information from removed data points.

Random relabeling is the online machine unlearning mechanism that sequentially retrains each to-be-deleted data point for $k'$ times (random $k'$ labels out of $k-1$ labels excluding target label) while each time the target label is replaced with a different random label. Since inaccurate data labels are introduced, a proper (un)learning rate for corresponding optimizer is selected to prevent severely contaminating the original model. Random relabeling is proposed based on the intuition that a random batch of complementary data points counters the influence of previously seen data point. Thus it is a less constraining mechanism of data removal. The removal is then certified by our proposed \textit{cosine divergence} metric that measures unlearning probability distribution similarity to naive approach. Once the maximum divergence, determined by membership inference, between two distributions is reached, model is retrained from scratch to fulfill all previous removal requests. Naive approach takes $\mathcal{O}(n)$ time to tackle each removal, while random relabeling is deletion efficient as it takes $\mathcal{O}(k')$ constant time in doing so. The main contributions of this paper are summarized as follows: 1) We propose the machine unlearning mechanism \textit{random relabeling} that sequentially fulfills each data deletion request in constant time; 2) A novel probability distribution metric based on cosine divergence and vector norm is developed as a less constraining data removal certification; 3) The unlearning algorithm consolidates well with original learning algorithm such that it can learn and unlearn data points alternatively with no conflicts.

\section{Literature Survey}
The term machine unlearning was first used in \cite{cao2015} for making statistical query learning systems forget. Recent discussions and regulations on data removal facilitate the study of machine unlearning in academia. Ginart \cite{ginart} investigated data deletion in a specific setting of $k$-means clustering, and \cite{guo2019certified} claimed that removal from such non-parametric model is trivial. The certified data removal mechanism from \cite{guo2019certified} only supports data removal from models with a convex loss function for calculating inverse Hessian matrix. However, the mechanism only applies to few models like logistic regressor, but not neural network-based models, due to loss function convexity restriction. Baumhauer \cite{baumhauer} proposed the linear filtration method for logit-based classifiers based on assumption of data removal for certain whole class. But such data removal for a entire class is not practically frequent. The SISA framework developed by \cite{sisa} divides the whole model into several constituents, and only one specific model constituent is retrained for each data point removal request. The drawback of the slicing strategy is that it introduces an expensive storage overhead for storing parameters for each model constituent. Therefore, an efficient and generic mechanism is an necessity for fulfilling data deletion request.


\section{Random Relabeling}
The goal of machine unlearning is to get the resulting model that is indistinguishable from the model trained from scratch after \textit{Data deletion} of data point $x$. We denote dataset consisting $n$ data points as $D=\{x_1, x_2, ..., x_n\}$, and each data point has $d$ dimensional features $x_i \in \mathcal{R}^d$. Let $A$ be a learning algorithm that maps a dataset to a model in hypothesis space $\mathcal{H}$. The \textit{data deletion} operation is defined as $R_A(D, A(D), i)$, for all $D$ and $i$, such that $A(D_{-i})$ is with (approximately) equal distribution with $R_A(D, A(D), i)$. Data point deletion in our approach is implemented in an online learning fashion where specific training algorithm $A(D)$ updates model parameters upon each given removal request.

We only consider the case of random relabeling applying to classification task, either binary class or multicalss classification, with $k$ denoting number of classes. Random relabeling is formally defined as a data-removal mechanism $M$ that removes the influence of data point $(x, y_c)$ by adding $k-1$ training data points $(x, \Tilde{y}) = \{(x, y_1), (x, y_2), ..., (x, y_{c-1}),(x, y_{c+1}), ..., (x, y_k)\}$ in a batch fashion. The naive retraining algorithm $A(D_{-i})$ processes each data point at least once in time $\mathcal{O}(n)$, while random relabeling achieves deletion efficiency in constant time $\mathcal{O}(k)$.

The model accuracy will inevitably drop somewhat after introducing a batch of complementary data points, which especially hurts the inference accuracy for data samples with same class as removed data points. We show the initial experimental results with MNIST \cite{mnist}, where it has 60, 000 training samples, and 10,000 test samples. The training and unlearning algorithm is built on the same ConvNet architecture with 2 convolution layers with default adaptive learning rate method \cite{adadelta}.
As shown in Table \ref{mnist}, testing accuracy decreases only around 2\%, after sequential removal of 600 data points, and almost no performance drop is observed for sequential removal of 100 data points.

\begin{table*}
\fontsize{10pt}{10pt}
\centering
\caption{Machine unlearning results on MNIST using naive and random relabeling approach. Original method corresponds to the case of no data deletion request. Unlearning size indicates the length of removal sequence.}
\begin{tabular}{|c| c| c| c| c|}
 \hline
   Method & Unlearning size & Standard Testing Set & Unlearned Set & Total Testing Set \\
 
 \hline  
Original & 0 & 99.02\% & 99.00\% & 99.02\% \\

\hline 
Naive & 100 & 99.03\% & 99.00\% & 99.03\% \\
Our approach & 100 & 99.02\% & 99.00\% & 99.02\% \\
\hline 
Naive & 600 & 99.14\% & 99.00\% & 99.13\% \\
Our approach & 600 & 96.99\% & 97.67\% & 97.03\% \\
 
 \hline
 \end{tabular}
\label{mnist}
\end{table*}

\begin{figure*}
\centering
\includegraphics[width=16cm]{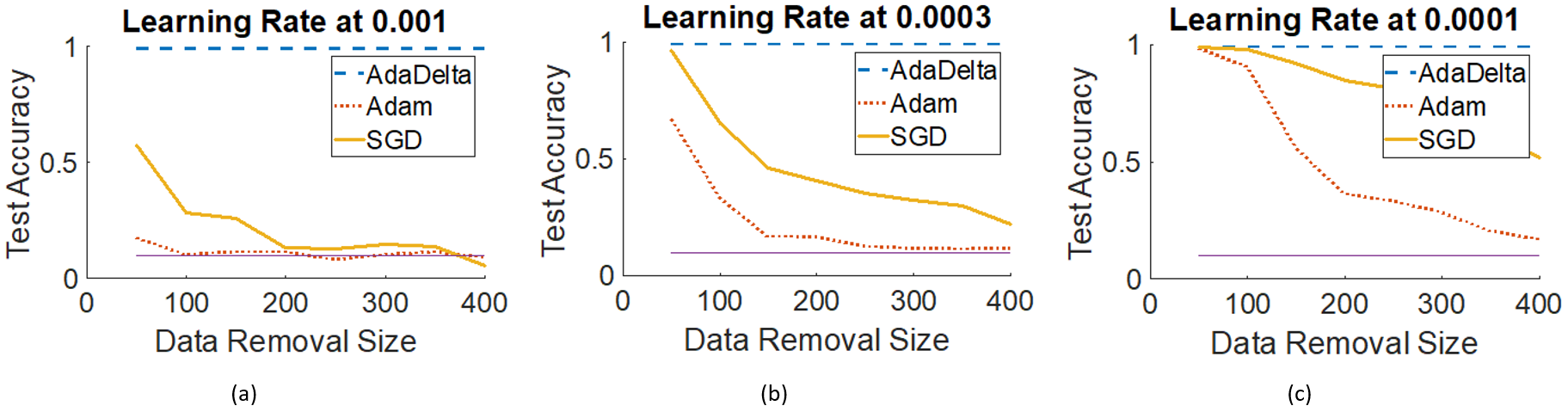}
\caption{Performance drop trend for different optimizers across three learning (unlearning) rates. Experiments are implemented on sequential removal requests from size of 50 to 400 at granularity of 50. The straight line shows random guess rate at 10\% for MNIST.}
\label{lr}
\end{figure*}

Considering the performance drop after a long stream of data deletion requests, two possible methods are proposed here to improve inference accuracy: 1) determining a suitable sequence length $l$ of removal requests, depending on acceptable accuracy standard of specific application scenario. Whenever the sequence length $l$ is reached, the model is retrained from scratch for accurately accommodating these $l$ removal request. For instance, $l=600$, 1 percent of full training set of MNIST, might be a good choice for setting up retraining point. 2) retraining only few redundant training data points upon each data deletion request. These redundant data points can be chosen from data set with same class as the removed data point, however, it brings time complexity up to $\mathcal{O}(n/k)$. Retraining on partial redundant data set compensates the performance drop well.

The effectiveness of removing impact of certain data point using random relabeling is self-explanatory, since the influence of previously seen is countered by introducing batch of complementary data points. Though certified removal is not the main focus of this work, the quality of unlearning if preferably proved by distribution divergence between $A(D_{-i})$ and $R_A(D, A(D), i)$. After extensive search of current literature, distribution divergence can be properly measured by metrics based on Bayes error rate \cite{guo2019certified} and differential privacy \cite{dwork2014algorithmic}.

\section{Tasks}
Following two tasks need to be properly handled before random relabeling is best implemented for efficient machine unlearning. One task is data removal certification, and another one is online unlearning configuration.

\textbf{Removal Certification.} We propose to solve this task from two perspectives. On one hand, data removal using random relabeling is sufficiently self-explanatory. Machine learning algorithm $A$ recognizes the pattern of training samples through parameter updates based on loss function for each single or batch of data points. Previously seen data point $(x, y_c)$ influenced the learning algorithm by updating model parameters during training session, however the batch of complementary data points $(x, \Tilde{y})$ apparently counteracts the influence of previous $(x, y_c)$. Moreover, recently fed data points have a more immediate effect on learning model update than previous data points. On the other hand, as mentioned above, data deletion can be certified using existing metrics such as Bayes error rate and differential privacy, or other metric that satisfies less constraining notion of data removal. Unlearning by random relabeling can be approximately certified with a method close to differential privacy as follows:

\begin{equation}\label{e1}
   P(R_A(D, A(D), i) \in \mathcal{T}) \leq \delta P(A(D_{-i}) \in \mathcal{T})
\end{equation}

where measurable subset $\forall \mathcal{T} \in \mathcal{H}$ and removed data point $\forall i \in D$. The definition states that the probability that data deletion operation using complementary relabeling returns a model from all possible hypothesis set and all removed samples, cannot be greater than $\delta^{-1}$ of the corresponding probability of the model returned from algorithm $A(D_{-i})$ \cite{ginart}.

\textbf{Online Unlearning Configuration.} There are several caveats to the online unlearning configuration when sequential removal requests are dealt with. Upon each removal request, these $k-1$ complementary data points should be fed into the model in a single mini-batch, in order to prevent more severe accuracy drop due to larger number of parameter updates. Optimizer and learning (or unlearning) rate also play important roles in inference accuracy. Their effects are investigated with three optimizers, i.e. AdaDelta \cite{adadelta}, Adam \cite{adam}, and SGD with three different learning rates, 0.001, 0.0003, and 0.0001. As shown in Figure \ref{lr}, AdaDelta is the most robust optimizer across data removal size from 50 to 400, and no performance drop is observed. Accuracy drops more significantly with larger learning rate. SGD behaves with fair accuracy only at small removal size, while accuracy with Adam optimizer fastly decreases to the level of random guess for all learning rates. Refer to \cite{adadelta} for advantages of AdaDelta. Therefore, proper sequence length $l$ mentioned in Section 3 is set based on considerations of optimizer, learning rate, and application scenario.

\section{Conclusion}

We proposed the novel idea of random relabeling for efficient machine unlearning. The approach approximately removes the influence of previously seen data points and requires no retraining which is computationally expensive and time consuming. The unlearning effect was validated on MNIST dataset with different optimizer and learning rate configurations. Differential privacy is required for removal certification which is not analyzed though.








{\small
\bibliographystyle{ieee_fullname}
\bibliography{egbib}
}

\end{document}